\documentclass[sigconf,screen]{acmart}  

\AtBeginDocument{%
  \fancypagestyle{plain}{%
    \fancyhf{} %
    \fancyfoot[L]{ACM SIGENERGY Energy Informatics Review}%
\fancyfoot[R]{Volume 4 Issue 4, October 2024}
  \providecommand\BibTeX{{%
    \normalfont B\kern-0.5em{\scshape i\kern-0.25em b}\kern-0.8em\TeX}}}
    
\let\oldmaketitle\maketitle
\renewcommand{\maketitle}{%
  \oldmaketitle%
  \thispagestyle{plain}%
  \pagestyle{plain}}
}

\usepackage{graphicx,subcaption,lipsum}
\usepackage{adjustbox}
\numberwithin{equation}{section}
\newcommand{\bs}{\boldsymbol}
\newcommand{\isdef}{\mathrel{\mathrel{\mathop:}=}}

\def\letters{a,b,c,d,e,f,g,h,i,j,k,l,m,n,o,p,q,r,s,t,u,v,w,x,y,z}
\def\Letters{A,B,C,D,E,F,G,H,I,J,K,L,M,N,O,P,Q,R,S,T,U,V,W,X,Y,Z}
\makeatletter
\@for \@l:=\Letters \do{%
  \expandafter\edef\csname\@l bb\endcsname{\noexpand\ensuremath{%
  \noexpand\mathbb{\@l}}}%
  \expandafter\edef\csname\@l bf\endcsname{{\noexpand\bf \@l}}%
  \expandafter\edef\csname\@l cal\endcsname{\noexpand\ensuremath{%
  \noexpand\mathcal{\@l}}}%
  \expandafter\edef\csname\@l eu\endcsname{\noexpand\ensuremath{%
  \noexpand\EuScript{\@l}}}%
  \expandafter\edef\csname\@l frak\endcsname{\noexpand\ensuremath{%
  \noexpand\mathfrak{\@l}}}%
  \expandafter\edef\csname\@l rm\endcsname{{\noexpand\rm \@l}}%
  \expandafter\edef\csname\@l scr\endcsname{\noexpand\ensuremath{%
  \noexpand\mathscr{\@l}}}%
}
\@for \@l:=\letters \do{%
  \expandafter\edef\csname\@l bf\endcsname{{\noexpand\bf \@l}}%
  \expandafter\edef\csname\@l frak\endcsname{\noexpand\ensuremath{%
  \noexpand\mathfrak{\@l}}}%
  \expandafter\edef\csname\@l scr\endcsname{\noexpand\ensuremath{%
  \noexpand\mathscr{\@l}}}%
} 

\begin{document}
\title{Probabilistic energy forecasting through quantile regression in reproducing kernel Hilbert spaces}

\author{Luca Pernigo}
\email{luca.pernigo@usi.ch}
\affiliation{%
  \institution{Euler Institute, USI}
  \streetaddress{Via la Santa 1, 6962}
  \city{Lugano}
  \country{Switzerland}}

\author{Rohan Sen}
\email{rohan.sen@usi.ch}
\affiliation{%
  \institution{Euler Institute, USI}
  \streetaddress{Via la Santa 1, 6962}
  \city{Lugano}
  \country{Switzerland}}

\author{Davide Baroli}
\email{davide.baroli@usi.ch}
\affiliation{%
  \institution{Euler Institute, USI}
  \streetaddress{Via la Santa 1, 6962}
  \city{Lugano}
  \country{Switzerland}}

\graphicspath{{./}}
\begin{abstract} 
Accurate energy demand forecasting is crucial for sustainable and resilient energy development. To meet the Net Zero Representative Concentration Pathways (RCP)  $4.5$ scenario in the DACH countries, increased renewable energy production, energy storage, and reduced commercial building consumption are needed. This scenario's success depends on hydroelectric capacity and climatic factors. Informed decisions require quantifying uncertainty in forecasts. This study explores a nonparametric method based on \emph{reproducing kernel Hilbert spaces (RKHS)}, known as kernel quantile regression, for energy prediction. Our experiments demonstrate its reliability and sharpness, and we benchmark it against state-of-the-art methods in load and price forecasting for the DACH region. We offer our implementation in conjunction with additional scripts to ensure the reproducibility of our research.

\end{abstract}

\ccsdesc[500]{Mathematics of computing~Nonparametric statistics}
\ccsdesc[500]{Computing methodologies~Feature selection}
\ccsdesc[500]{Software and its engineering~Software libraries and repositories}

\keywords{Kernel method, quantile regression, energy forecast, probabilistic forecast, climate,  GEFCom, SP2050, DACH}

\bibliographystyle{ACM-Reference-Format}

\maketitle

\section{Introduction}
Climate shock and the penetration of renewable energy sources are pivotal issues in the modern energy system, particularly in the DACH region (comprising Germany, Austria, and Switzerland).   Recently, the Swiss Federal Office of Energy has published a comprehensive analysis to assess how to secure and produce a cost-efficient energy supply in \href{https://www.bfe.admin.ch/bfe/en/home/policy/energy-strategy-2050.html}{Swiss Energy Strategy 2050} and \href{https://www.bfe.admin.ch/bfe/en/home/policy/energy-perspectives-2050-plus.html/}{Energy Perspectives 2050+}. Recently, Switzerland also adopted the long-term goal of climate neutrality, aiming to decrease its energy-building consumption and deploy more renewable energy technologies.  In addition to being essential to mitigate climate change, the performance of renewable energy sources and the demand for building energy depend on weather data and energy storage. 

Within the SURE SWEET energy initiative, supported by the SFOE, the development of future sustainable and robust systems is corroborated by techno-economic models that predict long-term scenarios and pathways, which are resilient to climate shocks \cite{Panos2023}. These models require a large amount of technical and economic data and their quality influences the reliability of the results, for example, bottom-up techno-economics model \cite{Kannan2018}, which provides hourly prediction, EXPANSE \cite{Trutnevyte2013},  building stock model \cite{Ngeli2020}, macro-economics \href{https://ec.europa.eu/clima/sites/clima/files/strategies/analysis/models/docs/gem e3 long en.pd}{GEM-3M} and life cycle assessment \cite{Luh2023}. The projection of these models is affected by weather data. For example, the Swiss building stock model designs decarbonisation pathways for different buildings' archetypes, whose isolation and heating performance vary with external temperature.  As a result, the prediction of hourly loads by transmission system operators is influenced by the variability and uncertainty of climate factors and is dependent on many parameters of the techno-economics models.

In particular, the pathways predicted by SURE models to achieve the Swiss net-zero scenario in conjunction with the RCP pathway $4.5$ address the primary issue of reliable energy supply. This is essential because it requires an increase in renewable energy sources to meet annual net electricity demand of $80$–$100$ TWh by $2050$, compared to the current $60$ TWh \href{https://www.newsd.admin.ch/newsd/message/attachments/73021.pdf}{SFOE 2022b}. One of the factors of such an increase in electricity due to sustainable mobility is up to 22 TWh \cite{Panos2022}.  To guarantee uninterrupted energy supply, even under extreme weather conditions \cite{HoTran2024}, in \href{https://www.fedlex.admin.ch/eli/fga/2022/2403/de}{SFOE 2022a} various scenarios have been examined: dependence on importing electricity from the European market and expansion of technologies that can provide or save electricity in winter, for example wind, alpine photovoltaic, seasonal heat storage or nuclear power.  


In such energy scenarios, forecasting models are needed to provide reliable energy management and probabilistic projections of socio-economical energy technologies \cite{Zielonka2023}. In addition, these models serve as a decision support tool for the transmission system operator (TSO), such as SwisseGrid, to determine the balance of reserves \cite{Abbaspourtorbati2016} and for policy-makers to develop a transition to sustainable energy sources. Due to the challenges described above, this work aims to investigate probabilistic forecasting to assess the uncertainty of energy supply due to fluctuations in hydroelectric capacity at day frequency, meteorological, and the intermittency of renewable energy sources at high temporal frequency, i.e., at hour resolution. 

In electricity forecasting state-of-the-art research, the focus has been mostly on point-forecast methods, that is, methods that output a single value for each target timestamp. Point forecasts are usually assessed using well-known criteria such as the \emph{root mean squared error (RMSE)} and the \emph{mean absolute error (MAE)}. Lately, the electricity forecasting community is shifting towards the probabilistic forecasting framework. The advantage of these methods is that they are more informative than a single-point prediction. 
\cite{gneiting2007probabilistic} introduces how to assess the quality of probabilistic forecasts by maximizing the sharpness of prediction distributions under calibration constraints. Calibration refers to the statistical consistency between the predicted distributions and the observations, while sharpness refers to the spread of the forecast distributions. Scoring rules assign numerical scores to the forecasts based on the predicted distribution and the value that is materialized. In the assessment of the probabilistic model, the most widely used scores are pinball loss and the \emph{continuous ranked probability score (CRPS)}. \cite{GNEITING2011197} studies the class of loss functions that lead to optimal predictors for the quantiles of a predictive distribution.

Probabilistic forecasting is gradually becoming an active research area for academia, where different researchers propose parametric and non-parametric models for forecasting specific outputs: wind, solar \cite{gneiting2023probabilistic}, prices \cite{NOWOTARSKI20181548} or demand \cite{Phipps2023}. A tutorial review on probabilistic load forecasting by \cite{hong2016probabilistic} covers forecasting techniques, auxiliary methodologies, evaluation metrics, and good sources of reference. Another review paper by \cite{NOWOTARSKI20181548} presents measures, tests, and guidelines for the rigorous use of probabilistic electricity forecasting methods. Another study, see \cite{van2018review}, provides a broad overview of probabilistic forecasting, specifically, the authors focus on solar power and load forecasting. In \cite{ziel2018probabilistic} an extensive literature review has been carried out by classifying electricity price forecasting papers according to various attributes such as prediction horizon, data used, predictors, accuracy measures, and models proposed.
 
This article addresses probabilistic forecasting by adopting the \emph{kernel quantile regression (KQR)} method within the RKHS framework. This method was introduced in \cite{takeuchi2006nonparametric} and further investigated in \cite{Li2007, JMLR_zhang2016,Sangneir2016,Zheng2021}. The method offers a non-parametric and non-linear way to provide probabilistic forecasts. The main contribution of this article is to perform a probabilistic forecast with KQR for Swiss, Austrian, and German energy systems, where the data are extracted from \href{transparency.entsoe.eu/)}{ENTSO-E Transparency Platform}, SECURES-Met \cite{Formayer2023} and C3S Energy\cite{Dubus2023}, which is designed to assess the impacts of climate variability and climate change on the energy sector. The probabilistic forecast with KQR has also been validated in the GEFCom test case, where our Python-based open-source implementation compares favourably with the top teams in the probabilistic forecast of electricity load and price.
 
Kernel quantile regression has received little attention in the energy forecasting literature. \cite{he2018probability} employs it to forecast wind power generation and compares it to a quantile regression neural network. \cite{moreira2016probabilistic} uses it to forecast electricity prices for the Iberian electricity market. In that study, kernel quantile regression is compared with linear quantile regression, neural network quantile regression, random forest, and regression boosting. \cite{he2017short} studies the choice of kernel functions in the context of short-term load forecasting.
However, no research has yet been done that thoroughly compares KQR to other state-of-the-art methods in the specific context of medium-term electricity load forecasting. 
Therefore, our second contribution is applying kernel quantile regression to the medium load forecasting setting, see section \ref{GEFCom2014}, sticking to best practices and guidelines of popular literature reviews in the field of probabilistic electric load forecasting (PLF) \cite{lago, prob_elf, nowotarski}.
We implemented our version of KQR since there were no implemented Python packages available. We made our model class inherit from the scikit template classes \texttt{BaseEstimator} and \texttt{RegressorMixin}; by doing so, our KQR is compatible with useful scikit learn functionalities such as grid search, cross-validation, and metric scorers; that is, our method is compatible with the scikit-learn API, see \cite{JMLR_scikit}. Sharing our implementation with the research community is the last contribution of this paper.

\subsection{Outline}
The rest of this article is structured as follows. In Section \ref{sec:kernel_quantile_regression}, we review quantile regression, in particular, kernel quantile regression. In Section \ref{sec:simple_example}, KQR is benchmarked against other popular quantile regressor models on data extracted from \href{transparency.entsoe.eu/)}{ENTSO-E Transparency Platform}. Following, a kernel wise comparison is carried out on the \href{https://zenodo.org/records/7907883}{SECURES-Met} data \cite{Formayer2023}. Furthermore, we evaluate the performance of KQR in the context of the challenge GEFCom2014 in Section \ref{GEFCom2014}. Finally, in Section \ref{conclusion}, we conclude the article.
 
\section{Kernel quantile regression}\label{sec:kernel_quantile_regression}

We first briefly review quantile regression here and then cover the details regarding kernel quantile regression. \emph{Quantile regression (QR)} is a method used in various fields, such as econometrics, social sciences, and ecology, to analyse the empirical distribution. Estimates a target quantile of the response variable $y$ based on a predictor vector, $\bs x$. QR is more robust to outliers in the data compared to the usual least-squares regression. It is also suitable for cases where there is heteroscedasticity in the errors. Additionally, using a series of quantile values provides a better description of the entire distribution than a single value, such as the mean. \cite{koenker1978regression} showed that the \emph{pinball loss} function
\begin{equation}\label{eq:pinball loss}
\begin{aligned}
    \rho_q(u)=
    \begin{cases}
        qu \quad& \text{if   } \ u\geq 0, \\
        -(1-q)u \quad& \text{if   } \ u<0,
    \end{cases}
    \end{aligned}
\end{equation}
can recover a target quantile of interest, $q$ where $0\leq q \leq 1$. We refer the reader to Appendix~\ref{appendix:pinball_loss} for further details on the same.
\begin{figure}[!ht]
    \centering
    \includegraphics[width=0.8\linewidth]{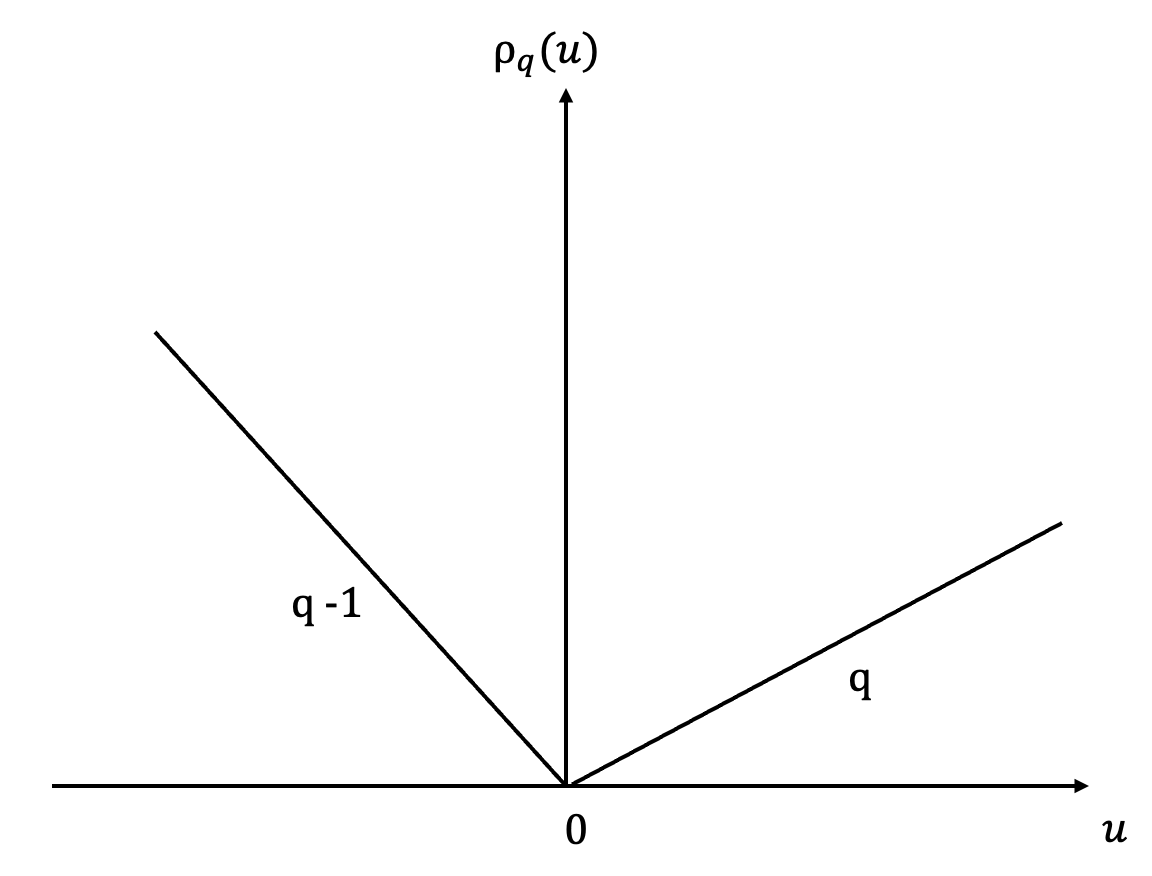}
    \caption[Pinball loss function at the $q$ quantile]{Pinball loss function at the $q$ quantile--\small \textmd{the lower the pinball loss, the more accurate the quantile forecast is.}}
    \Description{ }
    \label{fig:pinball_loss}
\end{figure}

Given empirical samples \((\bs x_i, y_i)_{i=1}^n\), with $\bs x_i \in \Rbb^d, \, y_i \in \Rbb$, quantile regression \cite{Koenker2001, koenker2005quantile} seeks to estimate the $q$-th conditional quantile of the response variable $y$ as a linear function of the explanatory variables $\bs x_i$ by solving the following minimization problem
\begin{equation}\label{eq:linear_quantile_regression}
    \underset{\bs \beta \in \Rbb^n} {\operatorname{argmin}}\sum_{i=1}^n \rho_q(y_i-\bs x_i^\top \bs \beta).
\end{equation}

With the motivation to perform QR that can also capture additional complexity and non-linearity in the data, several extensions of Problem~\ref{eq:linear_quantile_regression} were explored in \cite{Hwang2005, takeuchi2006nonparametric, Li2007, XU2015, JMLR_zhang2016}. In the above works, the authors investigated non-parametric versions of the QR, called the kernel quantile regression, which is based on the framework of reproducing kernel Hilbert spaces. The key idea of KQR is to first
transform the data samples non-linearly to a potentially higher-dimensional space of functions $\Hscr$ via a feature map $\phi(\cdot)$, and then consider an affine function of the transformed data.
Precisely, the conditional pinball loss of the response variable $\rho_q(y|\bs x)$ given the predictor vector $\bs x$ is approximated by functions of the form
\begin{equation}\label{eq:functional_form}
    f(\bs x) = \langle w, \phi(\bs x) \rangle_\Hscr + b,
\end{equation} 
where $w \in \Hscr$ is the regression coefficient and $b\in \Rbb$ is the intercept term. In particular, the map $x\mapsto \phi(\bs x)$ is implicitly defined by a reproducing kernel $\Kscr$ that makes $\Hscr$ an RKHS, which can contain a sufficiently rich class of functions. We first give herein the definition of RKHS and refer the reader to \cite{scholkopf2002learning, RKHS_book} for more details of the same.
\begin{definition}
Let $(\Hscr, \langle \cdot, \cdot \rangle_\Hscr)$ be a Hilbert space of real-valued functions on $\Xscr \subset \Rbb^d$. A function $\Kscr: \Xscr \times \Xscr \to \Rbb$ is called a \emph{reproducing kernel} of $\Hscr$ if and only if 
 \begin{align}\label{eq:rep_kernel}
    \Kscr(\bs x, \cdot) \in \Hscr &\quad \text{for all   } \bs x \in \Xscr;    \\
    \left\langle h, \Kscr(\bs x, \cdot)\right\rangle_\Hscr = h(\bs x) &\quad \text{for all   } h \in \Hscr, \, \bs x \in \Xscr. \label{eq:reproducing_property}
\end{align}
If the above two properties hold, $\Hscr$ is called a \emph{reproducing kernel Hilbert space}. Associated to every RKHS $\Hscr$, there exists the canonical \emph{feature map} \cite{aronszajn1950theory} $\phi: \Xscr \to \Hscr, \bs x \mapsto \Kscr(\bs x, \cdot)$ such that \begin{equation}
    \langle\phi(\bs x), \phi(\bs x') \rangle_\Hscr = \Kscr(\bs x, \bs x') \quad \text{for all   } \bs x, \, \bs x' \in \Xscr.
\end{equation} 
Furthermore, given a finite set of data samples $\{\bs x_1,\ldots, \bs x_n\} \subset \Xscr$, the \emph{kernel matrix} defined as \(\bs K \isdef [\Kscr(\bs x_i, \bs x_j)]_{i,j=1}^n \in \Rbb^{n \times n}\) is a symmetric and positive semi-definite matrix. 
\end{definition}

In the context of KQR, \cite{takeuchi2006nonparametric, Li2007, Zheng2021} consider the following regularised objective function 
\begin{equation}\label{eq:kqr1}
    R[f]:=\frac{1}{m}\sum\limits_{i=1}^{m}\rho_q(y_i-f(\bs x_i))+\frac{\lambda}{2}\|w\|_\Hscr^2,
\end{equation}
for $f(\bs x) = \langle w, \phi(\bs x)\rangle_\Hscr + b$, cp.\ \eqref{eq:functional_form}, where $w\in \Hscr$ and $\lambda > 0$ is a hyper-parameter. The first term in \eqref{eq:kqr1} measures the empirical loss in terms of the pinball function, cp.\ Equation~\ref{eq:pinball loss}, and the second term measures the complexity of the model, see \cite{vapnik1997support}. In particular, \cite{takeuchi2006nonparametric, Li2007, Zheng2021}  consider the following minimization problem
\begin{equation}\label{eq:kqr2}
     \underset{w \in \Hscr, b \in \Rbb}{\operatorname{argmin}} \, C \sum_{i=1}^n \rho_q\Big(y_i - \big(\langle w, \phi(\bs x_i)\rangle_\Hscr + b\big)\Big) + \frac{1}{2}\|w\|_\Hscr^2,
\end{equation}
where $C=\frac{1}{\lambda m} > 0$ is the factor that balances the model complexity and the total empirical pinball loss on the sample data. Using the representer theorem, see \cite{scholkopf2001generalized}, we have that the optimal solution to Problem~\ref{eq:kqr2} can be written as
a linear combination of kernel functions evaluated at the training examples, i.e.\ it has the following form
\begin{equation}\label{eq:optimal_functional_form}
    w^\star = \sum_{i=1}^n a_i^\star \phi(\bs x_i),
\end{equation}
or equivalently, 
\begin{equation}
   f^\star(\bs x) =  \sum_{i=1}^n a_i^\star \langle \phi(\bs x), \phi(\bs x_i) \rangle_\Hscr + b = \sum_{i=1}^n a_i^\star \Kscr(\bs x, \bs x_i) + b 
\end{equation}
The optimal coefficients $a_i^\star, \, 1\leq i\leq n$ are obtained via
\begin{equation}\label{eq:kqr_min6}
    \begin{aligned}
        \bs a^\star = \underset{\boldsymbol{a} \in \Rbb^n}{\operatorname{argmin}} \quad & \frac{1}{2} \boldsymbol{a}^\top \boldsymbol{K} \boldsymbol{a}-\boldsymbol{a}^\top \bs y\\
    \textrm{s.t.} \quad & 
    C(q-1)\textbf{1} \preceq \boldsymbol{a} \preceq Cq\textbf{1}\\
    & \boldsymbol{a}^\top\textbf{1}=0,
    \end{aligned}
    \end{equation}
    where $\bs y \isdef [y_i]_{i=1}^n \in \Rbb^n, \, \textbf {1} \isdef [1, \ldots 1]^\top \in \Rbb^n$, and $\bs K\in \Rbb^{n\times n}$ is the kernel matrix of the samples $(\bs x_i)_{i=1}^n$. We refer the reader to Appendix~\ref{appendix:kernel quantile regression derivation} for further details. Note that Problem~\ref{eq:kqr_min6} is a quadratic programming problem and, thus, can be solved by traditional solvers. 
\section{Numerical examples}\label{sec:simple_example}
All the numerical experiments have been performed in Python on a $3.2$ GHz $16$ GB Apple M1 Pro. Since KQR involves a quadratic programming problem, cp.\ ~\eqref{eq:kqr_min6}, we used the interior-point method implemented in the \texttt{cvxopt} library to solve it. For a detailed description of \texttt{cvxopt} solvers and algorithms available, we refer the interested reader to the manual \cite{vandenberghe2010cvxopt}. All scripts are made publicly available at the \href{https://github.com/luca-pernigo/kernel_quantile_regression}{Github repository} along with the cleaned data.

\subsection{Energy charts case study}
\begin{figure*}[!ht]
    \centering
    \includegraphics[width=\textwidth]{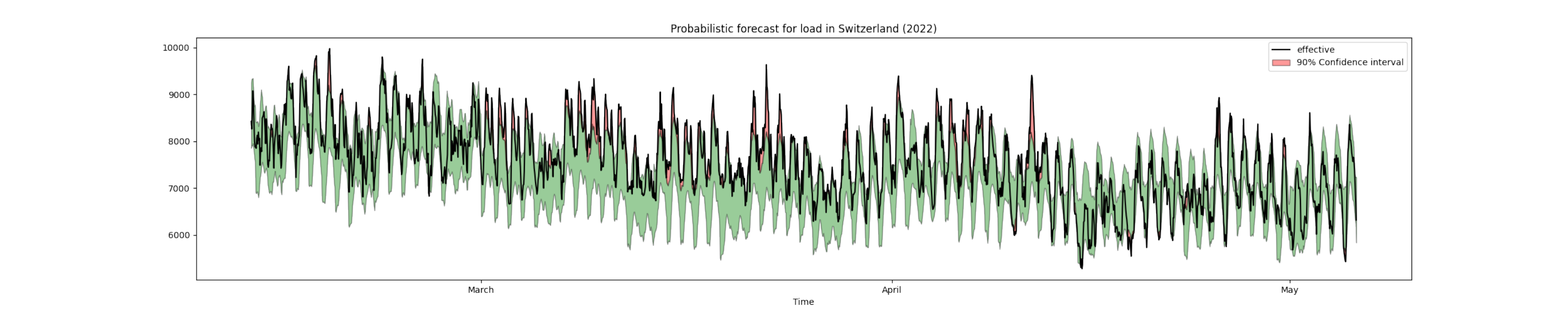}
    \Description{}
    \caption{Load 90\% confidence interval for Switzerland Energy charts data using KQR Absolute Laplacian: \small \textmd{Electric load probabilistic forecast for Switzerland 2022. The black line is the observed path for the load. The 90\% confidence interval bands are plotted in green. Lower and upper red lines denote the 95\% and 5\% quantile forecast respectively.}}
    \label{fig:CH_load_CI}
\end{figure*}

\begin{figure*}[!ht]
    \centering
    \includegraphics[width=\textwidth]{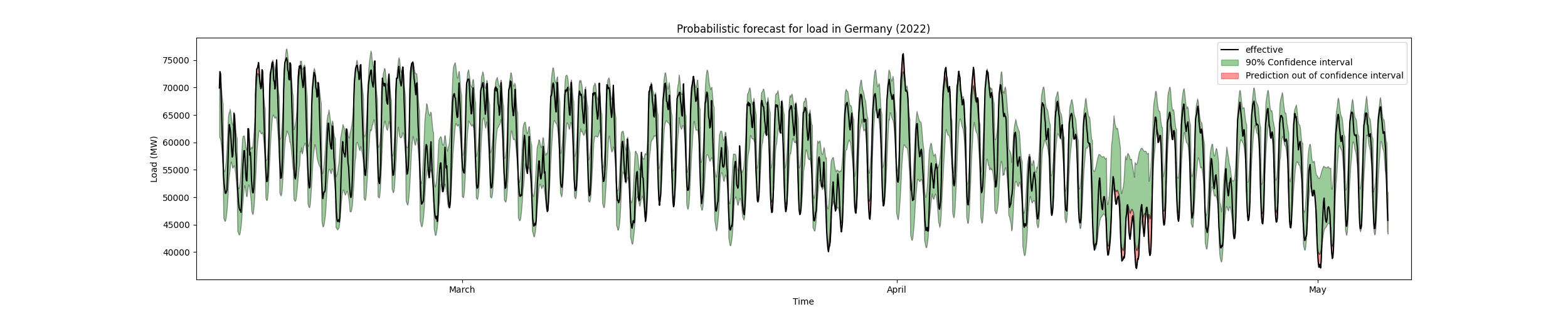}
    \Description{}
    \caption{Load 90\% confidence interval for Germany Energy charts data using KQR Absolute Laplacian: \small \textmd{Electric load probabilistic forecast Germany 2022. The black line is the observed path for the load. The 90\% confidence interval bands are plotted in green. Lower and upper red lines denote the 95\% and 5\% quantile forecast respectively.}}
    \label{fig:DE_load_CI}
\end{figure*}

In this case study, KQR is compared against popular quantile regressor models, the kernel of choice here is the Laplacian equipped with the Manhattan distance (Absolute Laplacian). The dataset for this case study comes from \href{https://www.energy-charts.info/index.html?l=en&c=DE}{Energy charts} which retrieves data from the \href{transparency.entsoe.eu/)}{ENTSO-E Transparency Platform}.
In predicting the national load we selected the following variables.
\begin{itemize}
    \item \texttt{Weather temperature};
    \item \texttt{Wind speed};
    \item \texttt{Hour};
    \item \texttt{Month};
    \item \texttt{Is holiday}: a binary variable for holidays where holiday=$1$, working day=$0$;
    \item \texttt{Day of week}: an ordinal categorical variable corresponding to the day of the week, e.i.\ Monday=$0$, \ldots Sunday=$6$;
   
\end{itemize}
We used the entire $2021$ data as the training sample for fitting our models, and we tested them and computed their scores on the $2022$ data. The results for Switzerland are reported in Table~\ref{tab:energy_chart_ch} and figure \ref{fig:CH_load_CI} while results for Germany can be found in Table~\ref{tab:energy_chart_de} and figure \ref{fig:DE_load_CI}.
The error metric used is the pinball loss, scaled by the average load magnitude of each country. It is observed that the KQR model performs marginally better than other QR models and achieves results comparable to those of the GBQR. The superior performance is demonstrated in terms of CRPS to its competitors, quantile regressors, in predicting electricity load quantiles.
\begin{table}[!ht]
\caption{Pinball loss for load in Switzerland (2022)}
\label{tab:energy_chart_ch}

\begin{tabular}{lrrrr}
\toprule
Quantile & LQR & GBMQR & QF & KQR\\
\midrule
0.1 & 0.03595 & 0.01243 & 0.01479 & \textbf{0.01210} \\
0.2 & 0.06161 & 0.01994 & 0.02200 & \textbf{0.01962} \\
0.3 & 0.08064 & 0.02573 & 0.02743 & \textbf{0.02495} \\
0.4 & 0.09396 & 0.02950 & 0.03073 & \textbf{0.02853} \\
0.5 & 0.10224 & 0.03174 & 0.03218 & \textbf{0.03048} \\
0.6 & 0.10393 & 0.03181 & 0.03115 & \textbf{0.03109} \\
0.7 & 0.09892 & 0.03009 & \textbf{0.02807} & 0.02958 \\
0.8 & 0.08528 & 0.02570 & \textbf{0.02293} & 0.02581 \\
0.9 & 0.05892 & 0.01862 & \textbf{0.01487} & 0.01845 \\
\midrule
CRPS & 0.08016 & 0.02506 & 0.02490 & \textbf{0.02451} \\
\bottomrule
\end{tabular}

\end{table}

\begin{table}[!ht]
\caption{Pinball loss for load in Germany (2022)}
\label{tab:energy_chart_de}

\begin{tabular}{lrrrr}
\toprule
Quantile & LQR & GBMQR & QF & KQR \\
\midrule
0.1 & 0.04051 & 0.02678 & \textbf{0.01585} & 0.01774 \\
0.2 & 0.06959 & 0.03020 & \textbf{0.02431} & 0.02517 \\
0.3 & 0.09069 & 0.03110 & 0.02852 & \textbf{0.02795} \\
0.4 & 0.10543 & 0.03011 & 0.03089 & \textbf{0.02881} \\
0.5 & 0.11379 & 0.02782 & 0.03200 & \textbf{0.02787} \\
0.6 & 0.11543 & \textbf{0.02485} & 0.02993 & 0.02558 \\
0.7 & 0.10934 & \textbf{0.02116} & 0.02610 & 0.02208 \\
0.8 & 0.09317 & \textbf{0.01650} & 0.02120 & 0.01737 \\
0.9 & 0.06319 & \textbf{0.01050} & 0.01264 & 0.01128 \\
\midrule
CRPS & 0.08901 & 0.02434 & 0.02460 & \textbf{0.02265} \\
\bottomrule
\end{tabular}

\end{table}

\subsection{SECURES-Met case study}
Combining the \href{https://zenodo.org/records/7907883}{SECURES-Met} data (predictors) \cite{Formayer2023} and the load data from \href{transparency.entsoe.eu/)}{ENTSOE}, we carried out a comparison between different classes of kernels. The kernel functions considered are: \href{https://scikit-learn.org/stable/modules/generated/sklearn.gaussian_process.kernels.RBF.html#sklearn.gaussian_process.kernels.RBF}{\texttt{Gaussian RBF}}, \href{https://scikit-learn.org/stable/modules/generated/sklearn.metrics.pairwise.laplacian_kernel.html#sklearn.metrics.pairwise.laplacian_kernel}{\texttt{Laplacian}}, \href{https://scikit-learn.org/stable/modules/generated/sklearn.gaussian_process.kernels.Matern.html#sklearn.gaussian_process.kernels.Matern}{\texttt{Matern} $0.5$, \texttt{Matern} $1.5$, \texttt{Matern} $2.5$}, \href{https://scikit-learn.org/stable/modules/generated/sklearn.metrics.pairwise.linear_kernel.html#sklearn.metrics.pairwise.linear_kernel}{\texttt{linear}}, \href{https://scikit-learn.org/stable/modules/generated/sklearn.gaussian_process.kernels.ExpSineSquared.html#sklearn.gaussian_process.kernels.ExpSineSquared}{\texttt{periodic}}, \href{https://scikit-learn.org/stable/modules/generated/sklearn.metrics.pairwise.polynomial_kernel.html#sklearn.metrics.pairwise.polynomial_kernel}{\texttt{polynomial}}, \href{https://scikit-learn.org/stable/modules/generated/sklearn.metrics.pairwise.sigmoid_kernel.html#sklearn.metrics.pairwise.sigmoid_kernel}{\texttt{sigmoid}}, and \href{https://scikit-learn.org/stable/modules/generated/sklearn.metrics.pairwise.cosine_similarity.html#sklearn.metrics.pairwise.cosine_similarity}{\texttt{cosine}}.
We used time series cross-validation to evaluate each model's performance, encompassing hyperparameter optimisation and feature selection. In Appendix~\ref{appendix:AppendixB}, Figure \ref{fig:GEFCOM_crossvalidation} shows the cross-validation process for the Absolute Laplace and Gaussian kernels. The analysis illustrates the criteria and scores to determine the optimal regularisation term and the hyperparameters of the kernel.

Details regarding the hyper-parameter selection for the different kernels can be found in the implementation \href{https://github.com/luca-pernigo/kernel_quantile_regression}{here}. The SECURES-Met data consist of historical data up to the end of $2020$ while from 2021 onward, the data consist of forecasts modelled by the European Centre for Medium-Range Weather Forecasts (ECMWF).
Therefore, we used the entire data of $2021$ as the training set and then tested our kernels on the $2022$ data.
Note that there are two types of prediction for the data from $2021$ onward, one for each of the emission scenarios RCP$4.5$ and RCP$8.5$. In this study, we restrict our attention to the RCP$4.5$ data, since it is part of SURE-SWEET's scenarios.

The predictors making up the dataset follow:
\begin{itemize}
    \item \texttt{Direct irradiation: direct normal irradiation.}
    \item \texttt{Global radiation: mean global radiation.}
    \item \texttt{Hydro reservoir: daily mean power from reservoir plants in MW.}
    \item \texttt{Hydro river: daily mean power from run of river plants in MW.}
    \item \texttt{Temperature: air temperature.}
    \item \texttt{Wind potential: potential wind power production.}
\end{itemize}
In Appendix~\ref{appendix:AppendixB}, we have reported the table of quantiles and CRPS scores. Table~\ref{tab:secures_met_ch} shows the results for Switzerland, Table~\ref{tab:secures_met_de} for Germany and Table~\ref{tab:secures_met_at} for Austria.


That study provides evidence of the superiority of the Gaussian kernel over the linear and polynomial kernels. In our research, we considered a larger set of kernels. 
From numerical experiments, the Absolute Laplacian kernel quantile has demonstrated superior performance to other Matern family kernels.  In addition, a comprehensive cross-validation process for quantile $0.5$ has been conducted to determine optimal hyperparameters and ridge regression parameters to prevent overfitting. The method has also been validated on an extended Secures Met dataset, which includes hydraulic capacity as a variable and technological-economic projections of the energy supply.
By including high-resolution hours as a categorical variable, we have achieved accurate predictions and narrower confidence bounds for the Absolute Laplacian and Gaussian RBF Kernels as shown in Figure \ref{fig:SECURES_Met_CH_a_laplacian} and Figure \ref{fig:SECURES_Met_CH_gaussian_rbf}. The superior performance of the Absolute Laplacian over the Gaussian RBF is largely attributable to the robustness of the Manhattan distance compared to the Euclidean distance\cite{Aggarwal2001}. The empirical demonstration is provided by comparing the covariance kernels in Figure \ref{fig:k_matrix_a_laplacian_gaussian_rbf_comp}. To complete our analysis,  Figures \ref{fig:CH_load_CI}  and \ref{fig:DE_load_CI} show a satisfactory prediction and narrow confidence for Switzerland and Germany even when the method slightly fails to approximate the real value.

\section{GEFCom2014 case study}\label{GEFCom2014}


We now apply KQR to the setting of probabilistic load and price forecasting. We use the GEFCom2014 \cite{hong2014global} data to carry out our experiments. The GEFCom is a series of competitions that have been created with the intent of improving forecasting practices, addressing the gap between academia and industry, and fostering state-of-the-art research in the field of energy forecasting \cite{hong2016probabilistic}. 
The GEFCom edition of $2014$ consisted of four tracks, all involving probabilistic forecasting. The four tracks were load, price, wind power, and solar power forecasting.
The reason for choosing the GEFCom2014 data is that it is an established benchmark in energy to compare against other valid methods. The data is freely available on Dr. Tao Hong's \href{http://blog.drhongtao.com/2017/03/gefcom2014-load-forecasting-data.html}{blog}.
Furthermore, a clear comparison can be performed due to the availability of the scores of each method for the load and price track. 
The score measure of the competition was the pinball loss, see section \ref{eq:pinball loss}, averaged over the $99$ quantiles, $q \in \{i/100\}_{i=1}^{n=99}$.
Finally, to recreate the setting of GEFCom2014 and to provide a fair comparison, we adhere rigorously to the rules of the competition. Next, we study the performance of KQR in the load and price tracks.
The kernel adopted throughout this study are the \href{https://scikit-learn.org/stable/modules/generated/sklearn.gaussian_process.kernels.RBF.html}{\texttt{Gaussian RBF and the Absolute Laplacian} kernel}.

\subsection{GEFCom2014 load track}\label{subsec:load experiment}
This track was concerned with forecasting hourly loads of an anonymous US utility. The dataset provided at the start of the competition consisted of $69$ months of load data and
$117$ months of weather data, both at hourly frequency.
In this particular track, the challenge was to predict the load for the next month without the availability of weather temperature forecasts. Therefore, the primary task was to first accurately predict the weather and temperatures and then model the load accordingly. Since there were no attributes available for humidity or wind speed, we chose to predict weather temperatures by aggregating historical temperature data across different dimensions such as day, month, and hour. Then we proceeded with building KQR models for the load; we chose the following predictors.
\begin{itemize}
    \item \texttt{Day: the number of the day}.
    \item \texttt{Hour}.
    \item \texttt{Day of week: an ordinal categorical variable for the day of the week}.
    \item \texttt{Is holiday: a dummy variable for holidays}. 
    \item \texttt{w avg: average of weather temperatures across all the $25$ stations}.
\end{itemize}
We built $12$ models, one for each month, with each task model trained on the historical data of the month associated with it.

Table \ref{tab:pinball loss gefcom2014 load data} reports our results for the load track. The top teams for the load forecasting track were Tololo, Adada, Jingrui(Rain) Xie, OxMAth, E.S. Managalova, Ziel Florian, and Bidong Liu; for a breakdown of the attributes of each method, see \cite[Table 6]{hong2016probabilistic}.

\begin{figure*}[!ht]
    \centering
    \includegraphics[width=\textwidth]{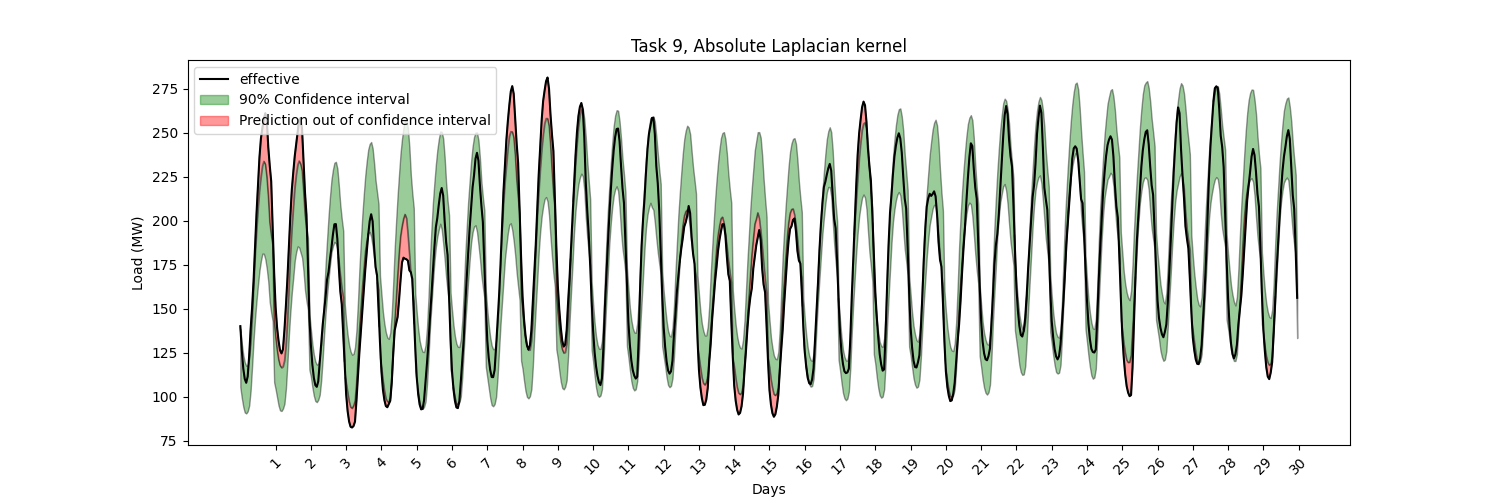}
    \Description{}
    \caption{Load 90\% confidence interval task 9 using KQR Absolute Laplacian: \small \textmd{Electric load probabilistic forecast for June 2011. The black line is the observed path for the load. The 90\% confidence interval bands are plotted in green. Lower and upper red lines denote the 95\% and 5\% quantile forecast respectively. The prediction out-of confidence interval is denoted in red.}}
    \label{fig:load_task_9}
\end{figure*}

We conclude this section with a visualisation of the $90\%$ confidence interval forecast by our model for task number $9$, that is the prediction for June, see Figure \ref{fig:load_task_9}.

The results demonstrate that kernel quantile regression yields scores similar to those of the top five methods outlined in \cite{he2017short}, specifically, the method has better performance with respect to non-quantile regression approaches, justifying the selection of this investigation. 

\subsection{GEFCom2014 price track}\label{subsec:price experiment}

In this track, the objective was to forecast electricity prices for the next $24$ hours on a rolling basis. The dataset provided consisted of $2.5$ years of hourly prices and zonal and system load forecasts.
The predictors fed to our KQR models are:
\begin{itemize}
    \item \texttt{Day};
    \item \texttt{Hour};
    \item \texttt{Forecasted total load};
    \item \texttt{Forecasted zonal load}.
\end{itemize}
Like above, all models were trained on the historical data of the associated month.

\begin{figure}[!ht]
    \centering
    \includegraphics[width=\linewidth]{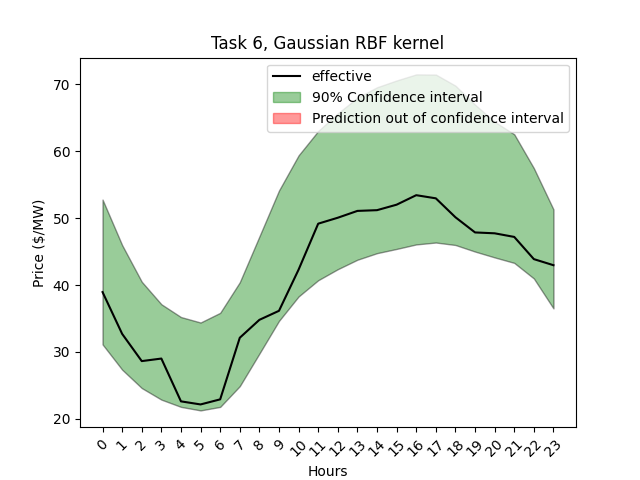}
    \caption{Price 90\% confidence interval task 6: \small \textmd{Electricity price probabilistic forecast for the $13^{\text{th}}$ July $2013$. The black line is the observed path for the price. The $90\%$ confidence interval bands are plotted in green. Lower and upper red lines denote the $95\%$ and $5\%$ quantile forecast respectively.}}
    \Description{}
    \label{fig:price_task_6_gaussian_rbf}
\end{figure}

Our results for this track are reported in table \ref{tab:pinball loss gefcom2014 price data}. In this track, the top entries came from the teams: Tololo, Team Poland, GMD, and C3 Green Team; for a breakdown of each method's attributes, see \cite[Table 8]{hong2016probabilistic}. Finally, our probabilistic prediction for the $13^{\text{th}}$ July $2013$ zonal price at the $90\%$ confidence interval is visualised in Figure \ref{fig:price_task_6_gaussian_rbf}.

\begin{table*}[!ht]
  \caption{Pinball loss GEFCom2014 load data}
  \label{tab:pinball loss gefcom2014 load data}
  \begin{adjustbox}{width=\textwidth}
  \begin{tabular}{lllllllllllll}
    \toprule
    \midrule
    Team name\textbackslash Task number                       & 1                               & 2                                  & 3                               & 4                              & 5                              & 6       & 7                               & 8       & 9       & 10                             & 11                             & 12                              \\
{KQR Absolute Laplacian}
& 
{12.5357}
&
{11.0209}
&
{9.4492}
&
{5.2240}
&
{6.6145}
&
{6.5418}
&
{11.2004}
&
{11.6325}
&
{5.9476}
&
{5.2219}
&
{7.5478}
&
{11.0587}
\\
{KQR Gaussian RBF}
& 
{12.4660}
&
{11.0894}
&
{9.4938}
&
{5.1826}
&
{6.9575}
&
{6.7947}
&
{10.8825}
&
{11.5542}
&
{5.9742}
&
{5.0779}
&
{7.3797}
&
{10.3110}
\\
\\
Adada                      & 10.5093                         & 10.0801                         & \textbf{7.6238}                          & 4.7289                         & \textbf{5.3936}                         & 6.6242  & 8.0144                          & 11.1366 & 5.7779  & 3.6379                         & 7.0096                         & 8.9109                          \\
Benchmark - Load           & 18.7384                         & 22.7585                         & 13.2163                         & 8.3626                         & 10.9162                        & 16.9937 & 13.4038                         & 17.3151 & 13.8374 & 6.4237                         & 10.9380                        & 34.0685                         \\
C3 Green Team              & 18.7384 & 19.2208 & 7.9637                          & 4.6370                         & 6.4543                         & 8.3799  & 10.5546                         & 10.6609 & 5.8867  & 4.4866                         & 5.9396                         & 10.3917                         \\
E.S. Mangalova             & 18.7384 & 13.3340 & 7.8025                          & \textbf{4.4096}                         & 6.6330                         & 6.2306  & 10.1511                         & 10.9294 & 6.2224  & 4.2382                         & 6.5464                         & 8.8080                          \\
Jingrui (Rain) Xie         & 11.8700                         & 10.9250                         & 8.4938                          & 4.9611                         & 7.4442                         & 6.9921  & 9.0523                          & 11.2600 & 5.4864  & \textbf{3.3602}                         & \textbf{5.9011}                         & 9.7316                          \\
OxMath                     & 14.4091 & \textbf{8.9136}                          & 7.6059                          & 4.4548                         & 7.2944                         & 7.4551  & 7.9527                          & \textbf{10.2444} & 5.4551  & 4.2111                         & 6.4054                         & 9.5520                          \\
Tololo                     & \textbf{10.4369}                         & 12.5232                         & 8.2695                          & 4.4220                         & 5.8976                         & \textbf{6.1878}  & \textbf{7.3182}                          & 10.8032 & \textbf{5.4469}  & 3.9613                         & 6.3173                         & \textbf{8.4787}                  \end{tabular}
\end{adjustbox}
\end{table*}

\begin{table*}[!ht]

  \caption{GEFCom2014 load data ranking}
  \label{tab:pinball loss gefcom2014 load data ranking}
  \begin{adjustbox}{width=\textwidth}
  \begin{tabular}{llllllllllllll}
    \toprule
    \midrule
    Team name\textbackslash Task number                       & 1                               & 2                                  & 3                               & 4                              & 5                              & 6       & 7                               & 8       & 9       & 10                             & 11                             & 12 & Aggregate ranking                              \\
KQR Absolute Laplacian
& 
6/362
&
7/362
&
14/362
&
10/362
&
4/362
&
3/362
&
20/362
&
15/362
&
9/362
&
16/362
&
18/362
&
12/362
&
11/362
\\
KQR Gaussian RBF
& 
5/362
&
8/362
&
15/362
&
9/362
&
8/362
&
7/362
&
18/362
&
12/362
&
10/362
&
15/362
&
17/362
&
8/362
&
10/362
\end{tabular}
\end{adjustbox}

\end{table*}

\begin{table*}[!ht]
  \caption{Pinball loss GEFCom2014 price data}
  \label{tab:pinball loss gefcom2014 price data}
  \begin{adjustbox}{width=\textwidth}
  \begin{tabular}{lllllllllllll}
    \toprule
    \midrule
    Team name\textbackslash Task number                       & 1                               & 2                                  & 3                               & 4                              & 5                              & 6       & 7                               & 8       & 9       & 10                             & 11                             & 12               \\
{KQR Absolute Laplacian}
&
{\textbf{1.02492}}
&
{3.35057}
&
{4.21374}
&
{7.33987}
&
{5.00981}
&
{6.96522}
&
{3.57168}
&
{1.77610}
&
{1.28765}
&
{2.73863}
&
{2.39831}
&
{23.30234}
\\
{KQR Gaussian RBF}
&
{1.84673}
&
{2.81882}
&
{1.54608}
&
{8.31636}
&
{4.05988}
&
{6.60456}
&
{3.57818}
&
{2.02177}
&
{1.45779}
&
{2.20701}
&
{1.98184}
&
{21.41033}
\\
\\

Benchmark - Price          & 4,02875          & 7,97208          & 5,70395          & 12,15104         & 38,33541         & 44,22979         & 18,22395         & 31,56729         & 42,94958         & 2,85583          & 3,20395          & 22,38333         \\
C3 Green Team              & 1,85897          & 3,27786          & 1,2593           & 5,08886          & 6,87674          & 6,1505           & 4,42379          & 1,32639          & 1,25915          & 3,08224          & 1,55811          & 6,58123          \\
GMD                        & 3,7271           & 1,783            & \textbf{0,92191}          & 5,08886          & 6,21331          & \textbf{3,82599} & 4,9342           & 1,47858          & 1,65933          & \textbf{2,06134}          & 2,1235           & 6,84571          \\
Team Poland                & 1,97477          & 1,81898          & 1,19162          & 2,82318          & 7,55914          & 4,20773          & 2,59715          & 1,04693          & \textbf{1,24193}          & 4,06012          & \textbf{1,08458} & \textbf{3,06512} \\
Tololo                     & 1,70734          & \textbf{1,45173} & 1,10384          & \textbf{2,01694} & 9,15596          & 4,6821           & \textbf{1,59517} & \textbf{0,75352}          & 2,45935          & 2,9614           & 1,34614          & 3,55819          \\
pat1                       & 2,36615          & 1,98567          & 1,07248          & 2,79465          & \textbf{4,23269} & 4,70614          & 8,40506          & 1,25376          & 2,23991          & 3,67952          & 1,06139          & 6,27517         

\end{tabular}
  \end{adjustbox}
\end{table*}

\begin{table*}[!ht]
  \caption{GEFCom2014 price data ranking}
  \label{tab:pinball loss gefcom2014 price data ranking}
  \begin{adjustbox}{width=\textwidth}
  
  \begin{tabular}{llllllllllllll}
    \toprule
    \midrule
    Team name\textbackslash Task number                       & 1                               & 2                                  & 3                               & 4                              & 5                              & 6       & 7                               & 8       & 9       & 10                             & 11                             & 12  & Aggregate ranking                            \\
KQR Absolute Laplacian
& 
1/287
&
11/287
&
12/287
&
14/287
&
4/287
&
11/287
&
5/287
&
8/287
&
6/287
&
8/287
&
15/287
&
18/287
&
15/287
\\
KQR Gaussian RBF
& 
6/287
&
8/287
&
10/287
&
15/287
&
1/287
&
10/287
&
6/287
&
10/287
&
7/287
&
5/287
&
9/287
&
16/287
&
10/287
\end{tabular}
}
\end{adjustbox}
\end{table*}

\section{CONCLUSION}\label{conclusion}
In this article, we investigate a non-parametric probabilistic method, the kernel quantile regression, for estimating quantiles of load. To our knowledge, this method has not been explored before for load prediction. We show its effectiveness through several numerical tests on DACH data (Germany, Austria and Switzerland), illustrating that it performs competently compared to other well-known quantile regression techniques and exceeds the point regression results of GEFCom2014.
In addition to these numerical experiments, we extend the test case of GEFCom2014 considering additional explanatory variables in hydrocapacity energy storage. We observe that KQR shows favourably and can forecast the medium-term horizon of the Secures-Met dataset. However, the forecasting for the short-term horizon has been demonstrated in the GEFCom2014 price track. The tuning of hyperparameters along with the ridge parameter was carried out using cross-validation, and we presented comprehensive analysis to support the results. In our investigation of kernel functions, we focus on evaluating the Absolute Laplace kernel in comparison to the RBF Gaussian Kernel during the kernel selection procedure. The results of this study confirm the durability of our selection in terms of precision of prediction. Finally, we provide an open-source implementation of KQR integrated with the most popular tools in the community. This investigation will serve as a case study for an uncertainty quantification model to predict the impact of climate shocks on pathways, which will be further refined in scenarios computed by the SURE SWEET energy models. Further investigation with different choices and combinations of kernels as well as a detailed comparison with gradient booster methods remains a future research area. 


\section{Acknowledgments}
We are grateful to three anonymous reviewers for insightful feedback. We are thankful for the fruitful discussions with Michael Multerer and Paul Schneider. Davide Baroli and Luca Pernigo carried out this research with the support of the Swiss Federal Office of Energy SFOE as part of the SWEET project SURE. The authors bear sole responsibility for the conclusions and the results presented in this publication. Rohan Sen was supported by the SNF grant ``Scenarios'' (100018\_189086).

\bibliography{sample-base}

\appendix
\section{Pinball Loss Function For Quantile Regression}\label{appendix:pinball_loss}
In this section, we show why minimizing the pinball loss function leads to estimates of the quantile. This section has been sourced from \cite{koenker2005quantile} and hence, we would like to refer to the above reference for further details on the same. For a real-valued random variable $X$ with distribution function $F(\cdot)$, the $q$-th quantile is defined as
\begin{equation} 
    Q(q) \isdef \operatorname{inf} \, \{x \in \Rbb: F(x) \ge q\} \quad \text{for $0\leq q \leq 1$}.
\end{equation}
For a continuous distribution function $F(\cdot)$, the quantile becomes just the inverse, i.e.\ $Q = F^{-1}$. For example, $q=0.5$ defines the median of the distribution of $X$. The quantiles arise from a simple optimization problem that is fundamental to
all that follows. Consider a simple decision-theoretic problem: a point estimate is required for a random variable with (posterior) distribution function $F$. If the loss is described by the piecewise linear pinball loss, cp.\ \eqref{eq:pinball loss}, consider the problem of finding $\hat{x}$ to minimize the expected loss. We seek to minimize
\begin{align}\label{eq:optimization_problem}
    \Ebb\left[\rho_q(X - \hat{x})\right] &= q \int_{\hat{x}}^{\infty} (x - \hat{x}) \, dF(x) \\ \nonumber
    &\, -(1-q)\int_{-\infty}^{\hat{x}} (x - \hat{x}) \, dF(x).
\end{align}
To find the optimal $\hat{x}$, we consider the first-order conditions by differentiating problem~\ref{eq:optimization_problem} with respect to $\hat{x}$ and setting it to zero:
\begin{align*}
    0 = - q \int_{\hat{x}}^{\infty} dF(x) + (1- q) \int_{-\infty}^{\hat{x}} dF(x) 
    = F(\hat{x}) - q
\end{align*}
Since $F(\cdot)$ is a monotone function, any element of $\{x: F(x)=q\}$ minimizes expected loss. When the solution is unique, $\hat{x} = F^{-1}(q)$, otherwise, we have an ``interval of $q$ quantiles'' from which the smallest element must be chosen - to adhere to
the convention that the empirical quantile function be left-continuous. 
When $F(\cdot)$ is replaced by the empirical distribution function 
\begin{equation}
    F_n(x) = \frac{1}{n} \sum_{i=1}^n \Ibb\{X_i \leq x\},
\end{equation}
we may still choose $\hat{x}$ to minimize expected loss:
\begin{equation}
    \int_{\Rbb} \rho_q(x - \hat{x}) \, dF_n(x) = \frac{1}{n} \sum_{i=1}^n \rho_q(x_i - \hat{x})
\end{equation}
where $x_i$'s are now assumed to be generated from the independently and identically distributed random variables $X_i \sim F(\cdot)$; doing so now yields the $q$-th sample quantile. When $qn$ is an integer there
is again some ambiguity in the solution, because we really have an interval
of solutions, $\{x : F_n(x) = q\}$, but this is of little practical consequence.

\section{Estimator derivation for kernel quantile regression}\label{appendix:kernel quantile regression derivation}
The goal is to solve the optimization problem, cp.\ ~\eqref{eq:kqr2}
\begin{equation}\label{eq:kqr_min1}
   \underset{w \in \Hscr, b \in \Rbb}{\operatorname{argmin}} \, C \sum_{i=1}^n \rho_q\Big(y_i - \big(\langle w, \phi(\bs x_i)\rangle_\Hscr + b\big)\Big) + \frac{1}{2}\|w\|_\Hscr^2, 
\end{equation}
with the characterization $f(\bs x) = \langle w, \phi(\bs x)\rangle_\Hscr + b$. Minimizing $\|w\|^2_{\Hscr}$ is equivalent to minimizing the regularizer, see \cite[Section 2.2.4]{scholkopf2002learning}. Using the definition of the pinball loss function, cp.\ \eqref{eq:pinball loss}, we have the problem 
\begin{equation}\label{eq:kqr_min2}
    \begin{aligned}
    \underset{w\in \Hscr, b \in \Rbb}{\operatorname{argmin}} \quad & C \sum \limits_{i=1}^{n}
    q(y_i-\langle w,\phi(\bs x_i) \rangle_{\Hscr}-b)+\\
    & +(1-q)(-y_i+\langle w,\phi(\bs x_i) \rangle_{\Hscr}+b)+ \frac{1}{2}\|w\|^2_{\Hscr},
    \end{aligned}
    \end{equation}
Introducing the slack variables $\xi_i, \, \xi_i^\ast, \, 1\leq i \leq n$, we can rephrase Problem~\ref{eq:kqr_min2} as the following constrained optimization problem, see \cite{Hwang2005, takeuchi2006nonparametric}
\begin{equation}\label{eq:kqr_min3}
    \begin{aligned}
        \underset{w\in\Hscr,\,b,\xi_i,\xi_i^\ast \in \Rbb}{\operatorname{argmin}} \quad & C \sum \limits_{i=1}^{m}
        q \xi_i+ (1-q)\xi_i^\ast+ \frac{1}{2}\|w\|^2_{\Hscr}\\
    \textrm{s.t.} \quad & y_i-\langle w,\phi(\bs x_i) \rangle_{\Hscr}-b \leq \xi_i,\\
    \quad & -y_i+\langle w,\phi(\bs x_i) \rangle_{\Hscr}+b \leq \xi_i^\ast,\\
      \quad & \xi_i\geq0,  \quad \xi_i^\ast\geq0. 
    \end{aligned}
    \end{equation}

Now, by the representer theorem, see \cite{scholkopf2001generalized}, any $w\in \Hscr$ that minimizes Problem~\ref{eq:kqr_min1} can be written as 
\begin{equation*}
    w = \sum_{i=1}^n a_i \phi(\bs x_i) \implies w(\bs x) = \sum_{i=1}^n a_i \Kscr(\bs x, \bs x_i).
\end{equation*}
Using \eqref{eq:reproducing_property}, we have that $\langle w, \phi(\bs x_j) \rangle_\Hscr = w(\bs x_j) = \sum_{i=1}^n a_i \Kscr(\bs x_i, \bs x_j)$ for $1 \leq j \leq n$. Hence, denoting the coefficient vector as $\bs a \isdef [a_i]_{i=1}^n \in \Rbb^n$, we have that 
\begin{equation}
    \left[\langle w, \phi(\bs x_j) \rangle_\Hscr\right]_{i=1}^n = \bs K \bs a, \qquad \|w\|_\Hscr^2 = \bs a^\top \bs K \bs a.
\end{equation}
Using the notations $\bs y \isdef [y_i]_{i=1}^n \in \Rbb^n, \bs \xi \isdef [\xi_i]_{i=1}^n \in \Rbb^n, \bs \xi^\ast \isdef [\xi_i^\ast]_{i=1}^n \in \Rbb^n$, we have the equivalent problem in matrix notation  
\begin{equation}\label{eq:kqr_min4}
    \begin{aligned}
        \underset{\bs a,\bs \xi, \bs \xi^\ast \in \Rbb^n, b \in \Rbb}{\operatorname{argmin}} \quad & C q \bs \xi^\top \mathbf{1}+ C (1-q) (\bs \xi^\ast)^\top \mathbf{1}+ \frac{1}{2}\bs a^\top \bs K \bs a\\
    \textrm{s.t.} \quad & \bs y-\bs K \bs a -b \mathbf{1} \preceq \bs \xi,\\
    & -\bs y+\bs K \bs a + b\mathbf{1} \preceq \bs \xi^\ast, \\
    &\bs \xi\succeq \bs 0, \quad \bs \xi^\ast\succeq \bs 0.
    \end{aligned}
    \end{equation}
Through the Lagrange multipliers and the KKT conditions (see \cite{boyd2004convex}), Problem~\ref{eq:kqr_min4} can be solved to its equivalent dual formulation, see \cite{takeuchi2006nonparametric, XU2015} for more details. We can write the Lagrangian as 
\begin{equation}\label{eq:kqr_min5}
    \begin{aligned}
    &L(\bs a, b, \bs \xi, \bs \xi^\ast, \bs \alpha, \bs \alpha^\ast, \bs \nu, \bs \nu^\ast) \isdef C q \bs \xi^\top \mathbf{1}+ C (1-q) (\bs \xi^\ast)^\top \mathbf{1}+ \frac{1}{2}\bs a^\top \bs K \bs a\\
    &- \bs \alpha^\top(\bs \xi - \bs y + \bs K\bs a + b\mathbf{1})
    - (\bs \alpha^\ast)^\top(\bs \xi^\ast + \bs y - \bs K\bs a - b\mathbf{1})\\
    & -\bs \nu^\top \bs \xi - (\bs \nu^\ast)^\top \bs \xi^\ast,
\end{aligned}
\end{equation}
with the positivity constraints $\bs \alpha, \bs \alpha^\ast, \bs \nu, \bs \nu^\ast \succeq \bs 0$. We proceed to derive the dual function by minimizing the Lagrangian
\begin{equation}\label{eq:dual_formulation}
   g(\bs \alpha, \bs \alpha^\ast, \bs \nu, \bs \nu^\ast)=\underset{\bs a, \bs \xi, \bs \xi^\ast \in \Rbb^n, b \in \Rbb}{\operatorname{inf}}
    L(\bs a, \bs \xi, \bs \xi^\ast, \bs \alpha, \bs \alpha^\ast, \bs \nu, \bs \nu^\ast).
\end{equation}
Setting the partial derivatives to zero, we have
\begin{equation}\label{eq:lagrange_derivatives}
    \begin{cases}
        \frac{\partial L}{\partial \bs a}= \bs 0 \implies \bs K \bs a=\bs K(\bs \alpha- \bs \alpha^\ast),\\
        \frac{\partial L}{\partial b}= 0 \implies (\bs\alpha-\bs \alpha^\ast)^\top\mathbf{1} = 0,\\
        \frac{\partial L}{\partial \bs \xi}= \bs 0 \implies Cq \mathbf{1}=\bs\alpha+ \bs \nu, \\
        \frac{\partial L}{\partial \bs \xi^\ast}= \bs 0 \implies C(1-q)\mathbf{1} = \bs \alpha^\ast +\bs \nu^\ast.
    \end{cases}
\end{equation}
Substituting the conditions into \ref{eq:kqr_min5}, we obtain the dual function as
\begin{equation}\label{eq:dual_problem}
    \begin{aligned}
        g(\bs \alpha, \bs \alpha^\ast, \bs \nu, \bs \nu^\ast) &= -\frac{1}{2}(\bs \alpha - \bs \alpha^\ast)^\top\bs K \bs (\bs \alpha - \bs \alpha^\ast)^\top 
        + (\bs \alpha - \bs \alpha^\ast)^\top\bs y,\\
         &\textrm{s.t.} \quad (\bs \alpha - \bs \alpha^\ast) = \bs a,\\
         & \qquad (\bs\alpha-\bs \alpha^\ast)^\top\mathbf{1} = 0,\\
         & \qquad \bs 0 \preceq \bs \alpha \preceq \bs Cq,\\
         & \qquad \bs 0 \preceq \bs \alpha^\ast \preceq \bs C(1-q).
    \end{aligned}
\end{equation}
In terms of the coefficient vector $\bs a$, we can consider the dual optimization problem which turns out to be (after switching signs)
\begin{equation}\label{eq:kqr_min7}
    \begin{aligned}
        \underset{\boldsymbol{a} \in \Rbb^n}{\operatorname{argmin}} \quad & +\frac{1}{2}\boldsymbol{a}^\top \bs K\boldsymbol{a}-\boldsymbol{a}^\top \bs y\\
    \textrm{s.t.} \quad & 
    C(q-1)\mathbf{1}\preceq \boldsymbol{a} \preceq Cq\mathbf{1}\\
    &\boldsymbol{a}^\top\mathbf{1}=0.
    \end{aligned}
    \end{equation}
From the constraint conditions of problem~\ref{eq:dual_problem} and the expression of $w$ cp.\ \eqref{eq:optimal_functional_form}, we have that the optimal $w$ is given by the coefficients
\begin{equation}
    w= \sum_{i=1}^n a_i^\star \phi(\bs x_i),
\end{equation}
where $\bs a^\star = [a_i^\star]_{i=1}^n$ is the solution of Problem~\ref{eq:kqr_min7}. Finally, the data points for which $a_i^\star \not \in \{C(q-1), Cq\}$ are called the support vectors. The intercept term $b$ can be calculated using the fact that $f(\bs x_i) = y_i$ for the set of support vectors, see \cite{takeuchi2006nonparametric, XU2015} for more details. The latter holds due to KKT conditions on Problem~\ref{eq:kqr_min7}.

\section{Additional numerical comparisons}\label{appendix:AppendixB}

In this Appendix, we present additional investigations that demonstrate the reliability and robustness of the KQR method introduced in the GEFCom2014 and Secures MET case study. 
\begin{table*}[!ht]
\caption{Pinball loss for load in Switzerland case study with SECURES-Met dataset}
\label{tab:secures_met_ch}
    \begin{adjustbox}{width=\textwidth}
        {
        \begin{tabular}{lllllllllll}
            \toprule
            Quantile & \begin{tabular}[c]{@{}l@{}}Absolute\\ Laplacian\end{tabular} & \begin{tabular}[c]{@{}l@{}}Matern 0.5/\\ Laplacian\end{tabular} & Matern 1.5 & Matern 2.5 & \begin{tabular}[c]{@{}l@{}}Matern $\infty$/\\ Gaussian RBF\end{tabular} & Linear & Periodic & Polynomial & Sigmoid & Cosine \\
            \midrule
            0.1 & \textbf{0.018783} & 0.018988 & 0.019108 & 0.019257 & 0.019547 & 0.019952 & 0.018891 & 0.019654 & 0.021456 & 0.021921 \\
            0.2 & \textbf{0.030441} & 0.030668 & 0.030809 & 0.031028 & 0.031546 & 0.032158 & 0.030683 & 0.031857 & 0.034032 & 0.034629 \\
            0.3 & \textbf{0.038467} & 0.038864 & 0.039050 & 0.039285 & 0.039907 & 0.040657 & 0.038968 & 0.040075 & 0.042944 & 0.043393 \\
            0.4 & \textbf{0.043622} & 0.044186 & 0.044516 & 0.044706 & 0.045286 & 0.046153 & 0.044354 & 0.045506 & 0.048701 & 0.048977 \\
            0.5 & \textbf{0.046160} & 0.046792 & 0.047205 & 0.047450 & 0.048116 & 0.048951 & 0.046956 & 0.047891 & 0.051423 & 0.051896 \\
            0.6 & \textbf{0.045499} & 0.046133 & 0.046824 & 0.047177 & 0.047840 & 0.048667 & 0.046446 & 0.047496 & 0.051292 & 0.051894 \\
            0.7 & \textbf{0.041494} & 0.042044 & 0.042928 & 0.043371 & 0.044104 & 0.044926 & 0.042458 & 0.043565 & 0.047656 & 0.048275 \\
            0.8 & \textbf{0.033837} & 0.034128 & 0.034797 & 0.035325 & 0.036166 & 0.037055 & 0.034290 & 0.035533 & 0.039501 & 0.040007 \\
            0.9 & 0.021883 & \textbf{0.021871} & 0.022431 & 0.022682 & 0.023169 & 0.023730 & 0.022000 & 0.022811 & 0.025561 & 0.026019 \\
            \midrule
            CRPS & \textbf{0.035576} & 0.035964 & 0.036407 & 0.036698 & 0.037298 & 0.038028 & 0.036116 & 0.037154 & 0.040285 & 0.040779 \\
            \bottomrule
    \end{tabular}
    }
\end{adjustbox}
\end{table*}

\begin{table*}[!ht]
\caption{Pinball loss for load in German case study with SECURES-Met dataset}
\label{tab:secures_met_de}
\begin{adjustbox}{width=\textwidth}
    {
    \begin{tabular}{lllllllllll}
        \toprule
        Quantile & \begin{tabular}[c]{@{}l@{}}Absolute\\ Laplacian\end{tabular} & \begin{tabular}[c]{@{}l@{}}Matern 0.5/\\ Laplacian\end{tabular} & Matern 1.5 & Matern 2.5 & \begin{tabular}[c]{@{}l@{}}Matern $\infty$/\\ Gaussian RBF\end{tabular} & Linear & Periodic & Polynomial & Sigmoid & Cosine \\
        \midrule
        0.1 & \textbf{0.025734} & 0.026272 & 0.026642 & 0.026845 & 0.027090 & 0.027306 & 0.026139 & 0.026621 & 0.027954 & 0.028004 \\
        0.2 & \textbf{0.043114} & 0.044138 & 0.044733 & 0.045057 & 0.045469 & 0.045917 & 0.044088 & 0.044716 & 0.046925 & 0.047025 \\
        0.3 & \textbf{0.054861} & 0.056343 & 0.056930 & 0.057335 & 0.057920 & 0.058528 & 0.056229 & 0.057066 & 0.060092 & 0.060209 \\
        0.4 & \textbf{0.061436} & 0.063490 & 0.064291 & 0.064748 & 0.065433 & 0.066136 & 0.063324 & 0.064683 & 0.067979 & 0.068202 \\
        0.5 & \textbf{0.064144} & 0.066330 & 0.067229 & 0.067696 & 0.068275 & 0.068960 & 0.066325 & 0.071421 & 0.070957 & 0.071359 \\
        0.6 & \textbf{0.062306} & 0.064676 & 0.065836 & 0.066299 & 0.066881 & 0.067405 & 0.064660 & 0.069160 & 0.068494 & 0.068780 \\
        0.7 & \textbf{0.055491} & 0.057851 & 0.058848 & 0.059317 & 0.059878 & 0.060275 & 0.061442 & 0.061423 & 0.060988 & 0.060990 \\
        0.8 & \textbf{0.044023} & 0.045011 & 0.045441 & 0.045644 & 0.046008 & 0.046374 & 0.047348 & 0.047182 & 0.047071 & 0.047056 \\
        0.9 & 0.026178 & \textbf{0.026126} & 0.026160 & 0.026187 & 0.026207 & 0.026282 & 0.026772 & 0.026766 & 0.026581 & 0.026628 \\
        \midrule
        CRPS & \textbf{0.048587} & 0.050026 & 0.050679 & 0.051014 & 0.051462 & 0.051909 & 0.050703 & 0.052115 & 0.053005 & 0.053139 \\
        \bottomrule
        \end{tabular}
    }
\end{adjustbox}
\end{table*}

\begin{table*}[!ht]
\caption{Pinball loss for load in Austrian case study with SECURES-Met dataset}
\label{tab:secures_met_at}
{
\begin{adjustbox}{width=\textwidth}
    \begin{tabular}{lllllllllll}
        \toprule
        Quantile & \begin{tabular}[c]{@{}l@{}}Absolute\\ Laplacian\end{tabular} & \begin{tabular}[c]{@{}l@{}}Matern 0.5/\\ Laplacian\end{tabular} & Matern 1.5 & Matern 2.5 & \begin{tabular}[c]{@{}l@{}}Matern $\infty$/\\ Gaussian RBF\end{tabular} & Linear & Periodic & Polynomial & Sigmoid & Cosine \\
        \midrule
        0.1 & \textbf{0.024365} & 0.024592 & 0.024550 & 0.024611 & 0.024831 & 0.026853 & 0.025744 & 0.026004 & 0.027672 & 0.027816 \\
        0.2 & \textbf{0.040971} & 0.041406 & 0.041413 & 0.041487 & 0.041857 & 0.045007 & 0.043247 & 0.043720 & 0.045899 & 0.045879 \\
        0.3 & \textbf{0.052316} & 0.052969 & 0.052888 & 0.052973 & 0.053449 & 0.057880 & 0.055231 & 0.055446 & 0.058572 & 0.058454 \\
        0.4 & \textbf{0.058286} & 0.059018 & 0.058916 & 0.059078 & 0.059592 & 0.065770 & 0.061844 & 0.062495 & 0.066599 & 0.066335 \\
        0.5 & \textbf{0.060344} & 0.061100 & 0.061319 & 0.061541 & 0.062338 & 0.068197 & 0.063864 & 0.071731 & 0.069761 & 0.069574 \\
        0.6 & \textbf{0.058724} & 0.059507 & 0.059674 & 0.059930 & 0.060709 & 0.065828 & 0.062265 & 0.070269 & 0.067179 & 0.066970 \\
        0.7 & \textbf{0.053148} & 0.054189 & 0.054249 & 0.054372 & 0.054916 & 0.058548 & 0.055830 & 0.063015 & 0.059807 & 0.059419 \\
        0.8 & \textbf{0.042740} & 0.043368 & 0.043408 & 0.043510 & 0.043881 & 0.046017 & 0.044384 & 0.049813 & 0.047078 & 0.046662 \\
        0.9 & \textbf{0.026430} & 0.026583 & 0.026781 & 0.026817 & 0.026908 & 0.027050 & 0.026471 & 0.030279 & 0.028031 & 0.027986 \\
        \midrule
        CRPS & \textbf{0.046369} & 0.046970 & 0.047022 & 0.047146 & 0.047609 & 0.051239 & 0.048764 & 0.052530 & 0.052289 & 0.052122 \\
        \bottomrule
    \end{tabular}
\end{adjustbox}
}
\end{table*}
\begin{figure*}[!ht]
    \centering
    \includegraphics[width=0.45\linewidth]{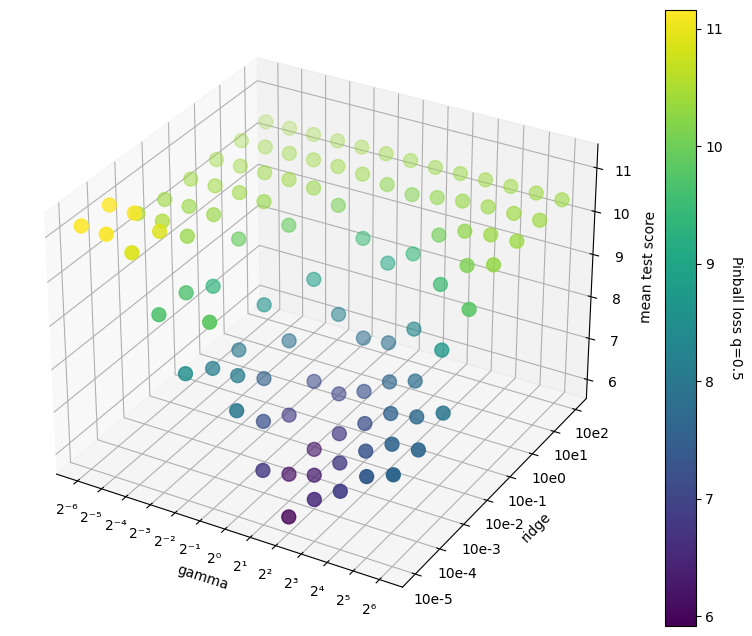}
       \includegraphics[draft=false, width=0.45\linewidth]{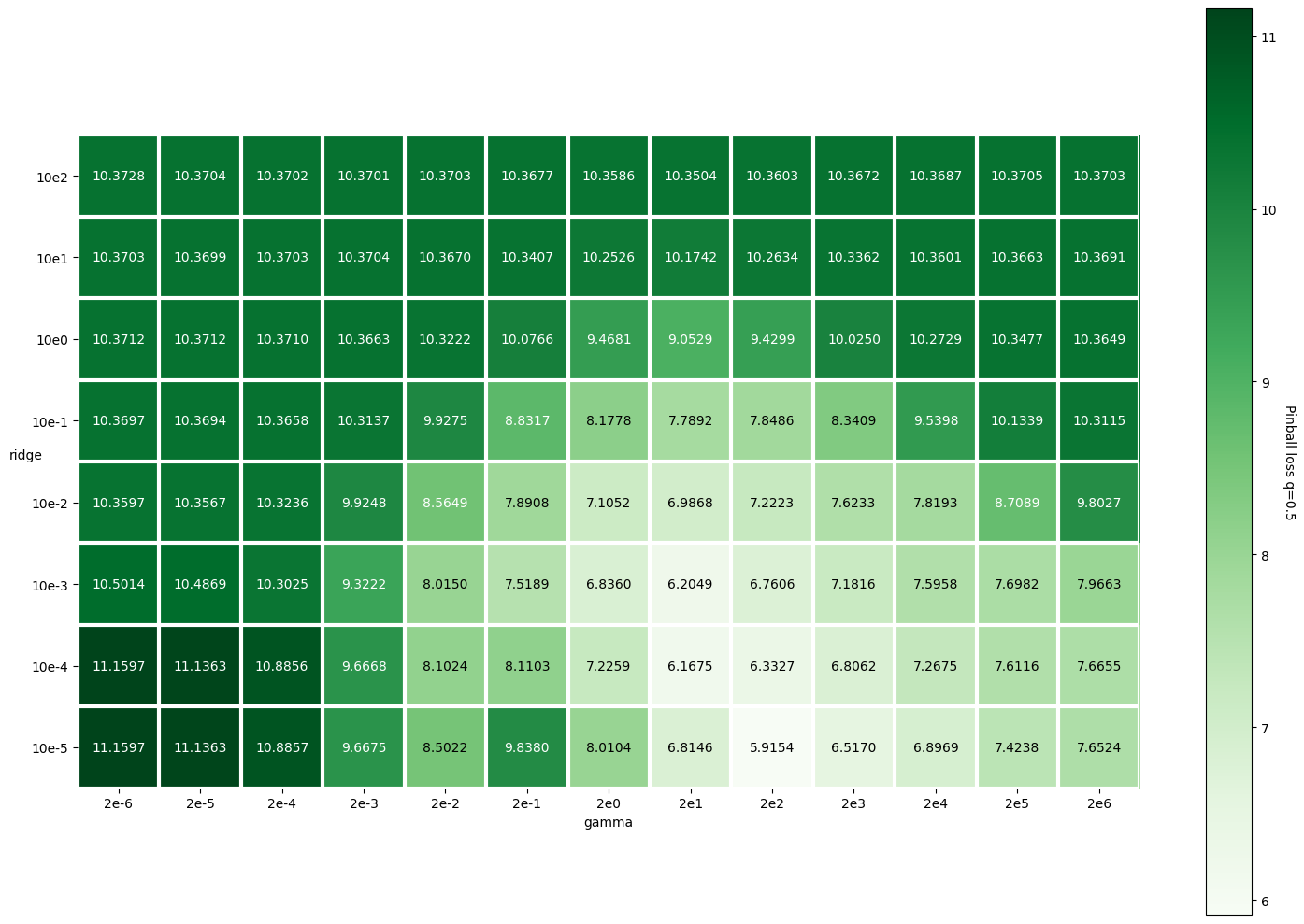}
        \caption{Hyperparameter cross-validation for the RBF Kernel Quantile Method applied to the price task 4, specifically the cross-validation is obtained for the median quantile. \small \textmd{On the left, the figure illustrates the optimal hyperparameters selection based on the mean quantile score. The same result is presented using a heatmap.}}
    \label{fig:GEFCOM_crossvalidation}
    \Description{}
\end{figure*}

\begin{figure*}[!ht]
    \centering
    \includegraphics[width=\linewidth]{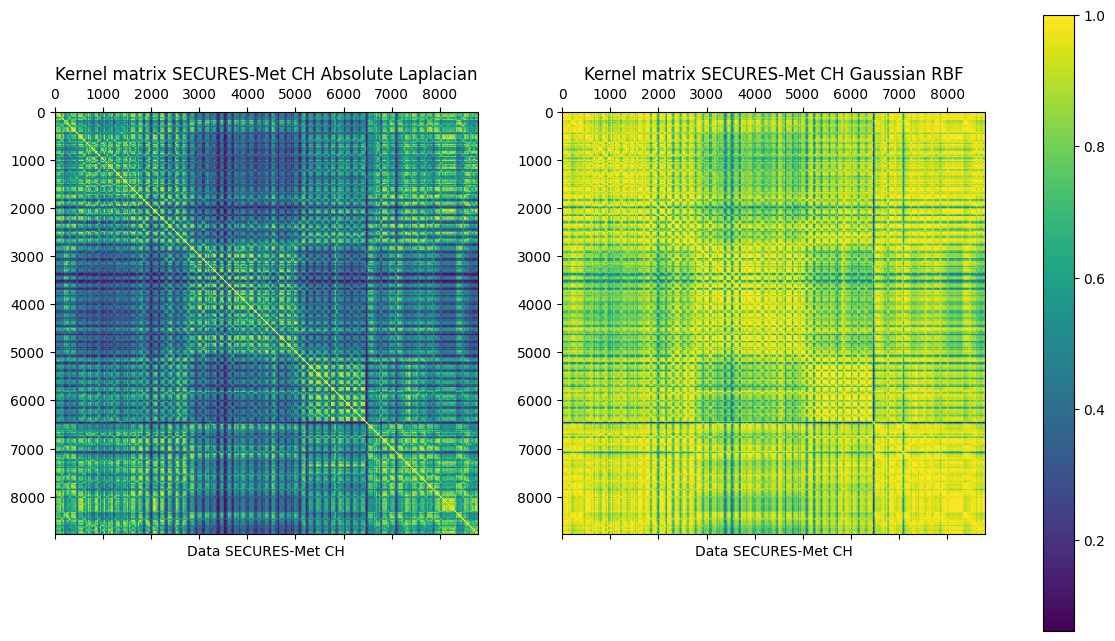}
    \Description{}
    \caption{Comparison between the kernel covariance evaluation for Absolute Laplacian and  Gaussian RBF kernel for SECURE-Met study.}
    \label{fig:k_matrix_a_laplacian_gaussian_rbf_comp}
\end{figure*}

\begin{figure*}[!ht]
    
    \includegraphics[draft=false,width=\linewidth]{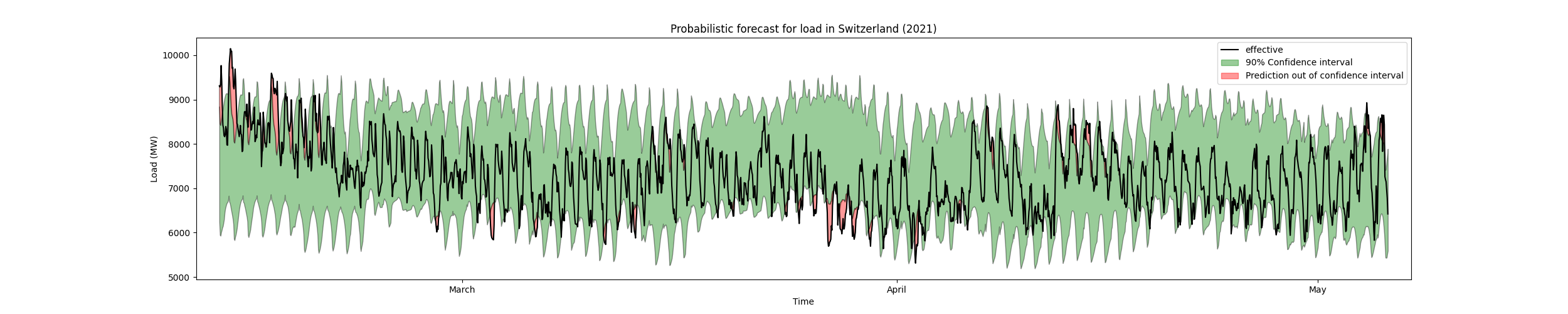}
    \Description{}
    \caption{Prediction and confidence bound of Absolute Laplacian kernel in Secures MET study. \small \textmd{The black line is the observed path for the load. The 90\% confidence interval bands are plotted in green. Lower and upper red lines denote the 95\% and 5\% quantile forecast respectively. The prediction out-of confidence interval is denoted in red.}}
    \label{fig:SECURES_Met_CH_a_laplacian}
\end{figure*}

\begin{figure*}[!ht]
    \centering
    \includegraphics[draft=false,width=\linewidth]{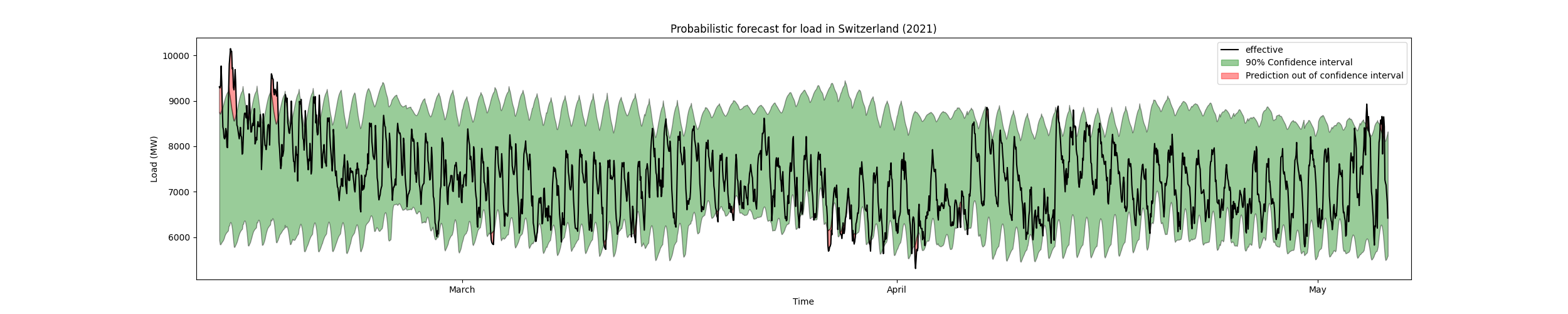}
    \Description{}
    \caption{Prediction and confidence bound of Gaussian RBF kernel in Secures MET study. \small \textmd{The black line is the observed path for the load. The 90\% confidence interval bands are plotted in green. Lower and upper red lines denote the 95\% and 5\% quantile forecast respectively. The prediction out-of confidence interval is denoted in red.}}
    \label{fig:SECURES_Met_CH_gaussian_rbf}
\end{figure*}

\end{document}